\documentclass[
    pre,
    twocolumn,
    twoside,
    showkeys,
    byrevtex,
    superscriptaddress,
    floatfix,
    nofootinbib,
    longbibliography
]{revtex4-1}

\usepackage[realmainfile]{currfile}
\input{\currfilebase.settings}
\usepackage{color}
\usepackage{xcolor}

\definecolor{skyblue}{HTML}{1569C7}
\definecolor{gray}{HTML}{808080}
\definecolor{firebrick}{HTML}{B22222}
\definecolor{seagreen}{HTML}{2E8B57}

\usepackage{url}
\PassOptionsToPackage{hyphens}{url}\usepackage{hyperref}
\makeatletter
\g@addto@macro{\UrlBreaks}{\UrlOrds}
\makeatother

\usepackage{hyperref}
\hypersetup{
  colorlinks=true,
  allcolors=skyblue,
  urlcolor=gray,
  citecolor=skyblue,
  pdfborder={0 0 0},
  breaklinks=true,
}

\usepackage{graphicx}
\usepackage{epsfig,verbatim,enumerate}

\usepackage{amssymb,amsmath}
\usepackage{ifthen}

\usepackage{tabularx}
\usepackage{multirow, makecell}
\newcolumntype{C}{>{\centering\arraybackslash}X}
\newcolumntype{L}{>{\raggedright\arraybackslash}X}
\newcolumntype{R}{>{\raggedleft\arraybackslash}X}
\usepackage{mathtools}

\newboolean{twocolswitch}

\newcommand{\sindex}[1]{}
\newcommand{\nindex}[1]{}

\newcommand{\www}[1]{\url{#1}}

\usepackage{lettrine}

\setboolean{twocolswitch}{true}

\begin{document}

\title{\protect
Augmenting semantic lexicons using word embeddings and transfer learning
}

\author{
\firstname{Thayer}
\surname{Alshaabi}
}
\email{thayer.alshaabi@uvm.edu}
 \affiliation{
  Advanced Bioimaging Center,
  UC Berkeley,
  Berkeley, CA.
  }
 \affiliation{
  Vermont Complex Systems Center,
  University of Vermont,
  Burlington, VT.
  }

\author{
\firstname{Colin M.}
\surname{Van Oort}
}
 \affiliation{
  The MITRE Corporation,
  McLean, VA.
  }
 \affiliation{
  Vermont Complex Systems Center,
  University of Vermont,
  Burlington, VT.
  }

\author{
\firstname{Mikaela Irene}
\surname{Fudolig}
}
\affiliation{
  Vermont Complex Systems Center,
  University of Vermont,
  Burlington, VT.
  }

\author{
  \firstname{Michael V.}
  \surname{Arnold}
}
\affiliation{
  Vermont Complex Systems Center,
  University of Vermont,
  Burlington, VT.
  }

\author{
  \firstname{Christopher M.}
  \surname{Danforth}
}
\affiliation{
  Vermont Complex Systems Center,
  University of Vermont,
  Burlington, VT.
  }
\affiliation{
  Department of Mathematics \& Statistics,
  University of Vermont,
  Burlington, VT.
  }

\author{
  \firstname{Peter Sheridan}
  \surname{Dodds}
}
\affiliation{
  Vermont Complex Systems Center,
  University of Vermont,
  Burlington, VT.
  }
 \affiliation{
  Department of Computer Science,
  University of Vermont,
  Burlington, VT.
  }

\date{\today}


\begin{abstract}
  \protect
  
Sentiment-aware intelligent systems are essential to a wide array of applications.
These systems are driven by language models
which broadly fall into two paradigms: 
Lexicon-based and contextual.
Although recent contextual models are 
increasingly dominant,
we still see demand for lexicon-based models 
because of their interpretability and ease of use.
For example, 
lexicon-based models allow
researchers to readily determine
which words and phrases
contribute most to a change in measured sentiment.
A challenge for any lexicon-based approach
is that the lexicon 
needs to be routinely expanded
with new words and expressions.
Here, 
we propose two models for automatic lexicon expansion.
Our first model establishes a baseline employing a simple and shallow neural network 
initialized with pre-trained word embeddings using a non-contextual approach.
Our second model improves upon our baseline, 
featuring a deep Transformer-based network 
that brings to bear word definitions to estimate their lexical polarity.
Our evaluation shows that both models are able to score new words 
with a similar accuracy to reviewers from Amazon Mechanical Turk, 
but at a fraction of the cost.

\end{abstract}



\maketitle


\section{Introduction}
\label{sec:sentiment.introduction}

In computational linguistics and natural language processing (NLP),
sentiment analysis involves extracting 
emotion and opinion from text data.  
There is an increasing demand for sentiment-aware intelligent systems.
The growth of sentiment-aware frameworks in online services
can be seen across a vast, multidisciplinary set of applications~\cite{nasukawa2003sentiment,medhat2014sentiment,bakshi2016opinion}.



With the modern volume of text data---which has long rendered human annotation infeasible---automated sentiment analysis is used, for example, by businesses in evaluating customer feedback to make informed decisions regarding 
product development and risk management~\cite{cabral2010dynamics,turney2002thumbs}. 
Combined with recommender systems, 
sentiment analysis has also been used 
with the intent to improve consumer experience 
through aggregated and curated feedback from other consumers, 
particularly in
retail~\cite{kumar2006retail,tang2009survey,yu2013impact},
e-commerce~\cite{bhatt2015amazon,haque2018sentiment},
and 
entertainment~\cite{terveen1997phoaks,pang2002thumbs}.

Beyond applications in industry, 
sentiment analysis has been widely applied in academic research, 
particularly in the social and political sciences~\cite{chen2021symbols}. 
Public opinion, e.g., support for or opposition to policies, can be potentially gauged from online political discourse, 
giving policymakers an important window into public awareness and attitude~\cite{laver2003extracting,thomas2006get}. 
Sentiment analysis tools have shown mixed results in forecasting elections~\cite{tumasjan2010predicting} and
monitoring inflammatory discourse on social media, 
with vital relevance to national security~\cite{pang2008opinion}. 
Sentiment analysis has also been used in the public health domain~\cite{coppersmith2014quantifying,yadollahi2017current,gohil2018sentiment}, 
with recent studies analyzing social media discourse surrounding mental health~\cite{bathina2021individuals,stupinski2021quantifying}, disaster response and emergency management~\cite{beigi2016overview}.


The growing number of applications of sentiment-aware systems has led the NLP community in the past decade to develop end-to-end models to examine short- and medium-length text documents~\cite{wilson2005recognizing,feldman2013techniques}, particularly for social media~\cite{pak2010twitter,agarwal2011sentiment,korkontzelos2016analysis}.
Some researchers have considered
the many social and political implications of using AI for sentiment detection across media~\cite{crawford2021excavating, crawford2019halt}.
Recent studies highlight some of the implicit hazards of crowdsourcing text data~\cite{shmueli2021beyond}, 
especially in light of the latest advances in NLP and emerging ethical concerns~\cite{hovy2016social, conway2016social}.
Identifying potential racial and gender disparity in NLP models is essential to develop better models~\cite{tatman2017gender}.


Sentiment analysis tools can be classified into two broad groups 
depending on their definition of sentiment and their model for its estimation. 
The probability of belonging to a discrete class 
(e.g., positive, negative)
is a common way of defining sentiment for a given piece of text.  
When edge cases are frequent, 
adding a neutral class has been reported to improve overall performance~\cite{ribeiro2016sentibench}. 
However, sometimes a cardinal measure is desired, 
requiring a spectrum of sentiment scores rather than a sentiment class~\cite{thelwall2010sentiment}. 
This more nuanced sentiment scoring paradigm has been widely adopted for e-commerce, movies, and restaurant reviews~\cite{snyder2007multiple}.

Sentiment analysis models largely derive from two major paradigms: 
1. Lexicon-based models 
and 
2. Contextual models.
Lexicon-based models compute sentiment scores based on sentiment dictionaries (sentiment lexicons) 
typically constructed by human annotators~\cite{taboada2011lexicon,dodds2015human,augustyniak2016comprehensive}. 
A sentiment lexicon contains not only terms that express a particular sentiment/emotion, 
but also terms that are associated with a particular sentiment/emotion (denotation vs. connotation).
Contextual models, on the other hand, 
extrapolate semantics by converting words to vectors in an embedding space, 
and learning from large-scale annotated datasets to predict sentiment based on co-occurrence relationships between words~\cite{wilson2005recognizing,pak2010twitter,agarwal2011sentiment,feldman2013techniques,socher2013recursive}. 
Contextual models have the advantage in differentiating multiple meanings, 
as in the case of ``The dog is \textit{lying} on the beach'' vs. ``I never said that---you are \textit{lying}'', 
while lexicon-based models usually have a single score for each word, regardless of usage.
Despite the flexibility of contextual models, 
their results can be difficult to interpret,
as the high-dimensional latent space in which they are embedded renders explanation difficult.
The ease of use and transparent comprehension of lexicon-based models help explain their continued popularity~\cite{pang2008opinion,taboada2011lexicon,dodds2015human}. 
For example,
while the linguistic mechanisms leading to change in sentiment may be hard to explain with word embeddings, 
one can straightforwardly use lexicon scores to reveal the words contributing to shifted sentiment~\cite{dodds2011temporal,reagan2017sentiment,gallagher2021generalized}.


A major challenge for the simpler and more interpretable lexicon-based models, 
however,
is the time and financial investment associated with maintaining them.
Sentiment lexicons must be updated regularly to mitigate the out-of-vocabulary
(OOV) problem---words and phrases that were either not considered or did not exist when the dictionaries were originally constructed~\cite{riloff1996empirical}.
While researchers show general sentiment trends are observable
unless the lexicon does not have enough words,
having a versatile dictionary with specialized and rarely used words improves the signal~\cite{dodds2010measuring,reagan2017sentiment}.
Notably,
language is an evolving sociotechnical phenomenon.
New words and phrases are created constantly,
especially on social media~\cite{alshaabi2021storywrangler}.
Words occasionally substitute others or drift in meaning over time.
For example,
the word `covid' grew to be the most narratively trending $n$-gram
in reference to the global Coronavirus outbreak during February and March 2020~\cite{alshaabi2021world}.


In this work,
we propose an automated framework extending sentiment for semantic lexicons to OOV words, 
reducing the need for crowdsourcing scores from human annotators, a process that can be time-consuming and expensive. 
Although our framework can be used in a more general sense, 
we focus on predicting \textit{happiness scores} based on the labMT dataset~\cite{dodds2015human}.
This dataset was constructed from human ratings of the ``happiness'' of words on a continuous scale, 
averaging scores from multiple annotators for more than 10,000 words. 
We discuss this dataset in detail in Sec.~\ref{subsec:sentiment.data}.
In Sec.~\ref{sec:sentiment.related_work},
we discuss recent developments using deep learning in NLP, and how they relate to our work.
We introduce two models,
demonstrating accuracy on par with human performance
(see Sec.~\ref{sec:sentiment.methods} for technical details).   
We first introduce a baseline
model---a neural network initialized with pre-trained word 
embeddings---to gauge happiness scores.
Second, 
we present a deep Transformer-based model that uses word definitions to estimate their sentiment scores.
We will refer to our models as the `Token' and `Dictionary' models, respectively.
We present our results and model evaluation in Sec.~\ref{sec:sentiment.results}, 
highlighting how the models perform compared with reviewers from Amazon's Mechanical Turk.
Finally,
we highlight key limitations of our approach, and outline some potential future developments in concluding remarks.

\section{Related work}
\label{sec:sentiment.related_work}


Word embeddings are abstract numerical representations of the relationships between words, 
derived from statistics on individual corpora, 
and encoding language patterns so that concepts with similar semantics have similar representations~\cite{bengio2003neural}. 
Researchers have shown that efficient representations of words can both 
express meanings and preserve context~\cite{maas2011learning,hollis2016principals,hollis2017extrapolating,li2017inferring}.
While there are many ways to construct word embedding models 
(e.g., matrix factorization), 
we often use the term to refer to a specific class of word embeddings that are learnable via neural networks.


Word2Vec is one of the key breakthroughs in NLP,
introducing an efficient way for learning word embeddings from a given text corpus~\cite{mikolov2013efficient,mikolov2013distributed}.  
At its core,
it builds off of a simple idea borrowed from linguistics and formally known as the
`distributional hypothesis'---words
that are semantically similar are also used in similar ways,
and likely to appear with similar context words ~\cite{harris1954distributional}.

Starting from a fixed vocabulary,
we can learn a vector representation
for each word via a shallow network with a single hidden layer trained in one of two fashions~\cite{mikolov2013efficient,mikolov2013distributed}.  
Both approaches formalize the task as a unsupervised prediction problem,
whereby an embedding is learned jointly with a network that is trained to
either predict an anchor word given the words around it
(i.e., continuous bag-of-words (CBOW)),
or by predicting context words for an anchor word
(i.e., skip-gram)~\cite{mikolov2013efficient}.
Both approaches, 
however, 
are limited to local context bounded by the size of the context window. 
Global Vectors (GloVe) addresses that problem 
by capturing corpus global statistics with a word co-occurrence probability matrix~\cite{pennington2014glove}.


While Word2Vec and GloVe offer substantial improvements over previous methods, 
they both fail to encode unfamiliar words---tokens that were not processed in the training corpora. 
FastText refines word embeddings by supplementing the learned embedding matrix with subwords 
to overcome the challenge of OOV tokens~\cite{joulin2017bag,bojanowski2017enriching}.
This is achieved by training the network with character-level $n$-grams ($n \in \{3, 4, 5, 6\}$), 
then taking the sum of all subwords to construct a vector representation for any given word. 
Although the idea behind FastText is rather simple, 
it presents an elegant solution to account for rare words, 
allowing the model to learn more general word representations.


A major shortcoming of the earlier models 
is their inability to capture contextual descriptions of words 
as they all produce a fixed vector representation for each word.
In building context-aware models, 
researchers often use fundamental building blocks such as 
recurrent neural networks (RNN)~\cite{rumelhart1986learning}---particularly 
long short-term memory (LSTM)~\cite{hochreiter1997long}---that 
are designed to process sequential data. 
Many methods have provided incremental improvements over time~\cite{peters2017semi,lee2017end,chen2017enhanced}.
ELMo is one of the key milestones towards efficient contextualized models, 
using deep bi-directional LSTM language representations~\cite{peters2018deep}.


In late 2017, 
the advent of Transformers~\cite{vaswani2017attention} 
rapidly changed the landscape in the NLP community.
The encoder-decoder framework, powered by attention blocks, 
enables faster processing of the input sequence while also preserving context~\cite{vaswani2017attention}.
Recent adaptations of the building blocks of Transformers continue to break records, 
improving the state-of-the-art across all NLP benchmarks 
with recent applications to computer vision and pattern recognition~\cite{dosovitskiy2021image}.


Exploiting the versatile nature of Transformers,
we observe the emergence of a new family of language models
widely known as ``self-supervised" including as
bidirectional encoders (e.g., BERT)~\cite{devlin2019bert},
and left-to-right decoders (e.g., GPT)~\cite{radford2018improving}.
Self-supervised language models are pre-trained
by masking random tokens in the unlabeled input data and training the model to predict these tokens.
Researchers leverage recent subword tokenization techniques, such as
WordPiece~\cite{wu2016wordpiece},
SentencePiece~\cite{kudo2018sentencepiece},
and Byte Pair Encoding (BPE)~\cite{sennrich2016neural},
to overcome the challenge of rare and OOV words.
Subtle contextualized representations of words can be learned
by predicting whether sentence B follows sentence A~\cite{devlin2019bert}.
Pre-trained language models can then be fine-tuned using labeled data for downstream NLP tasks,
such as
named entity recognition,
question answering,
text summarization,
and sentiment analysis~\cite{devlin2019bert,radford2018improving}.


Recent advances in NLP continue
to improve the language facility of Transformer-based models.  
The introduction of XLNet~\cite{yang2019xlnet} is another remarkable breakthrough that combines the
bi-directionality of BERT~\cite{devlin2019bert}
and the autoregressive pre-training scheme from Transformer-XL~\cite{dai2019transformer}.
While the current trend of making ever-larger and deeper language models shows an impressive track record,
it is arguably unfruitful to maintain unreasonably large models
that only giant corporations can afford to use due to hardware limitations~\cite{thompson2020computational}.  
Vitally, less expensive language models need to be both computationally efficient and exhibit performance on par with larger models.
Addressing that challenge,
researchers proposed clever techniques of leveraging knowledge distillation~\cite{distilling2015hinton}
to train smaller and faster models
(e.g, DistilBERT~\cite{sanh2019distilbert}).
Similarly,
efficient parameterization strategies via sharing weights across layers
can also reduce the size of the model while maintaining state-of-the-art results
(e.g., ALBERT~\cite{lan2020albert}).


Previous work on automatic sentiment lexicon generation (ASLG) has used a variety of heuristics to assign sentiment scores to OOV words.
Most ASLG methods start with a seed lexicon containing words of known sentiment,
then use a distance function to propagate sentiment scores from known words to unknown words.
Word co-occurrence frequencies~\cite{turney_Measuring_2003,kiritchenko_Sentiment_2014} and shortest path distances within a semantic word graph~\cite{qiu_Expanding_2009, baccianella_SentiWordNet_2010,sanvicente_Simple_2014} 
(such as WordNet~\cite{fellbaum_WordNet_1998}) 
were common distance functions in earlier work.
More recently, 
distance functions based on learned word embeddings have gained popularity~\cite{tang_Building_2014,wang_CommunityBased_2016,ljubesic_Predicting_2018,thavareesan_Sentiment_2020}. 
The outputs of word embedding models usually need to be projected into a lower dimension before they can be used for ASLG.
This can be done using a variety of machine learning models, though linear models are likely one of the most popular options~\cite{qiu_Expanding_2009,amir_INESCID_2015,wang_CommunityBased_2016,li_Inferring_2017,ljubesic_Predicting_2018,alshari_Effective_2018,thavareesan_Sentiment_2020}.
\citet{amir_INESCID_2015}~proposed the use of a support vector regressor (SVR) 
trained with 
CBOW~\cite{mikolov2013efficient} or 
GloVe~\cite{pennington2014glove} word embeddings, finding that the SVR model out performed various linear models 
(e.g., Lasso~\cite{yuan2006lasso}, Ridge~\cite{hoerl1970ridge}, ElasticNet~\cite{zou2005elastic} regressors) 
on the labMT lexicon.
However, 
their models only predicted a binary sentiment polarity ($\hat{y} \in [0, 1]$), 
rather than continuous scores. 
\citet{li_Inferring_2017}~extended their work, 
proposing a class of linear regression models trained with word embeddings to predict affective meanings in several sentiment lexicons such as ANEW~\cite{bradley1999affective}, VAD~\cite{mohammad2018obtaining}.
\citet{darwich_CorpusBased_2019}~present an excellent review of ASLG.

Building on the recent models discussed above, 
we develop a framework for augmenting semantic lexicons using word embeddings and pre-trained large language models.
Our models output continuous valued sentiment scores that can represent degrees of negative, neutral, and positive sentiment.
Our tool reduces the need for crowdsourcing scores from human annotators
while still providing similar,
and often better,
results compared with random reviewers from Amazon Mechanical Turk at a fraction of the cost.

\section{Data and methods}
\label{sec:sentiment.methods}

We propose two models for predicting happiness scores for the labMT
lexicon~\cite{dodds2015human}---a general-purpose sentiment lexicon used to measure happiness in text corpora
(see Sec.~\ref{subsec:sentiment.data} for more details).

Our first model is a neural network
initialized with pre-trained FastText word embeddings.
The model uses fixed word representations to gauge the happiness score for a given expression, enabling us to augment the labMT dataset at a low cost. 
For simplicity,
we will refer to this model as the Token model.

Bridging the link between lexicon-based and contextualized models, 
we also propose a deep Transformer-based model that uses word definitions
to estimate their happiness scores---namely, the Dictionary model.
The contextualized nature of the input data allows our model to accurately estimate the expressed happiness score for a given word based on its lexical meaning. 

We implement our models using
Tensorflow~\cite{abadi2016tensorflow}
and Transformers~\cite{wolf2020transformers}.
See
Sec.~\ref{subsec:sentiment.token_model}
and
Sec.~\ref{subsec:sentiment.dictionary_model}
for additional details of our Token and Dictionary models, respectively.
Our source code, along with pre-trained models, are publicly available via our GitLab repository
(\url{https://gitlab.com/compstorylab/sentiment-analysis}).

\subsection{Data}
\label{subsec:sentiment.data}

In this study, 
we use the labMT dataset as an example sentiment lexicon to test and evaluate our models~\cite{dodds2015human}.
The labMT lexicon contains roughly ten thousand unique
words---combining the five thousand most frequently used words from
New York Times articles,
Google Books,
Twitter messages,
and music lyrics~\cite{dodds2015human}.
It is a lexicon designed
to gauge changes in the happiness (i.e., valence or hedonic tone) of text corpora.
Happiness is defined on a continuous scale
$h \in \{1 \to 9\}$,
where $1$ bounds the most negative (sad) side of the spectrum,
and $9$ is the most positive (happy).
Ratings for each word are crowdsourced via Amazon Mechanical Turk (AMT),
taking the average score
$h_{avg}$
from 50 reviewers to set a happiness score for any given word.
For example, 
the words `suicide', `terrorist', and `coronavirus' have the lowest happiness scores, 
while the words `laughter', `happiness', and `love' have the highest scores.   
Function and stop words along with numbers and names tend to have neutral scores 
($h_{avg} \approx 5$), 
such as `the', `fourth', 'where', and `per'.

The labMT dataset also powers the Hedonometer,
an instrument quantifying daily happiness on Twitter~\cite{dodds2011temporal}.
Over the past few years,
the labMT lexicon was updated to include new words that were not found in the original survey
(e.g, terms related to the COVID19 pandemic~\cite{alshaabi2021world}).

We are particularly interested in this dataset
because it also provides the standard deviation of human ratings for each word,
which we use to evaluate our models.   
In this work,
we propose two models to estimate
$h_{avg}$
using word embeddings,
and thus provide an automated tool to augment the labMT dataset both reliably and efficiently.

\begin{figure}[!tp]
    \centering
    \includegraphics[width=\columnwidth]{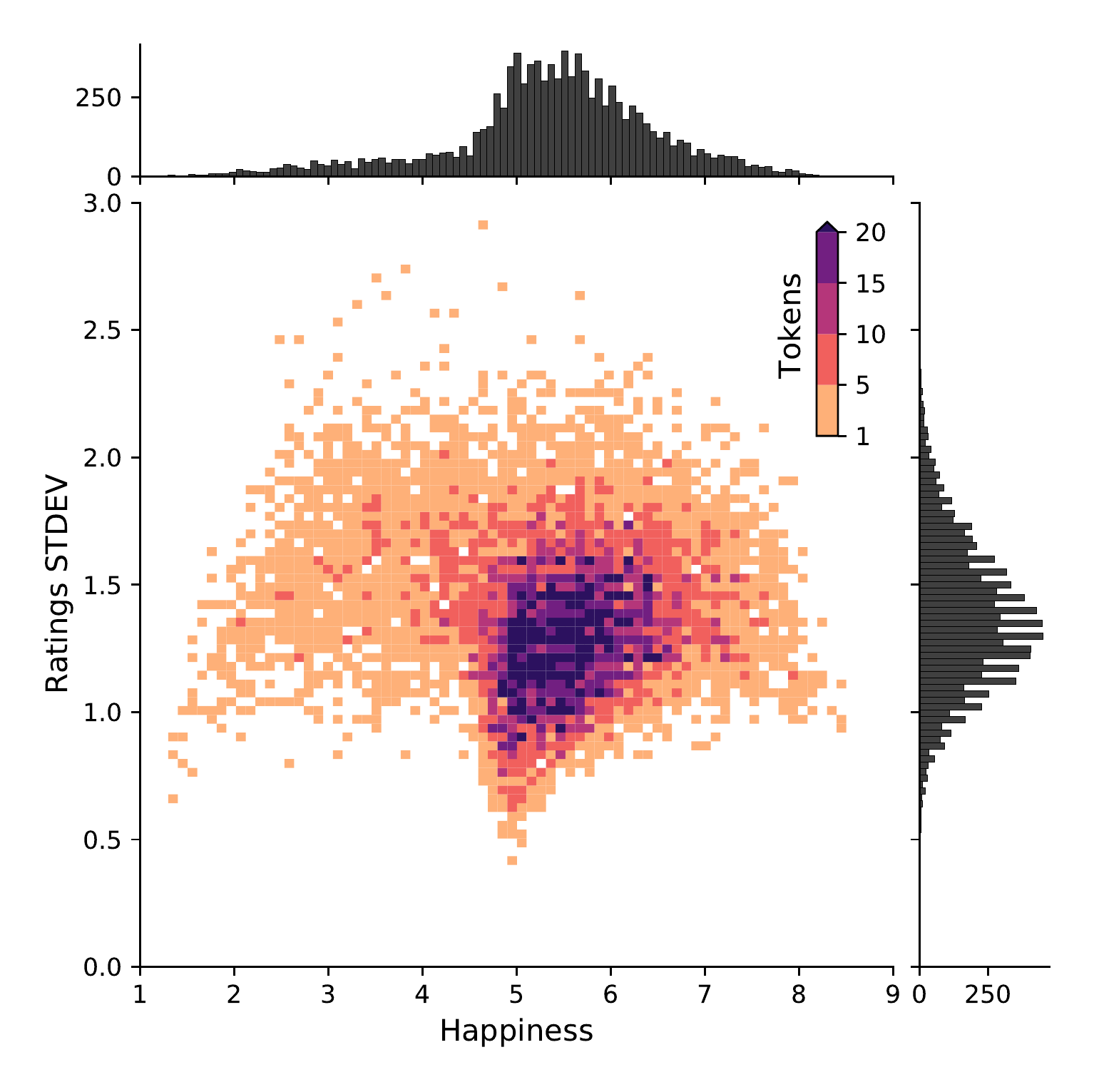}
    \caption{
    \textbf{Emotional valence of words and uncertainty in human ratings of lexical polarity.}
    A 2D histogram of happiness 
    $h_{avg}$ 
    and standard deviation of human ratings 
    for each word in the labMT dataset. 
    Happiness is defined on a continuous scale from $1$ to $9$, 
    where $1$ is the least happy and $9$ is the most.
    Words with a score between $4$ and $6$ are considered neutral.
    While the vast majority of words are neutral, 
    there is a positive bias in human language~\cite{dodds2015human}. 
    The average standard deviation of human ratings 
    for estimating the emotional valence of words in the labMT dataset is $1.38$.
    }
    \label{fig:happiness_heatmap}
\end{figure}

In Fig.~\ref{fig:happiness_heatmap},
we display a 2D histogram of the human rated happiness scores in the labMT dataset.
The figure highlights the degree of uncertainty in human ratings of the emotional valence of words.
For example, the word `the' has an average happiness score of $h_{avg} = 4.98$, with standard deviation of $\sigma = 0.91$, while the word `hahaha' has a happier score with $h_{avg} = 7.94$ and $\sigma = 1.56$.   
Some words also have a relatively large standard deviation such as 
`church' ($h_{avg} = 5.48, \sigma = 1.85$), 
and `cigarettes' ($h_{avg} = 3.31, \sigma = 2.6$).

While the majority of words are neutral, with a score between 4 and 6,
we still observe a human positivity bias in the English language~\cite{dodds2015human,mahabhaleshwara2021positivity}.
On average,
the standard deviation of human ratings is $1.38$.
In our evaluation (Sec.~\ref{sec:sentiment.results}),
we show how our models perform relative to the uncertainty observed in human ratings.

\begin{figure*}[!tp]
    \centering
    \includegraphics[width=\textwidth]{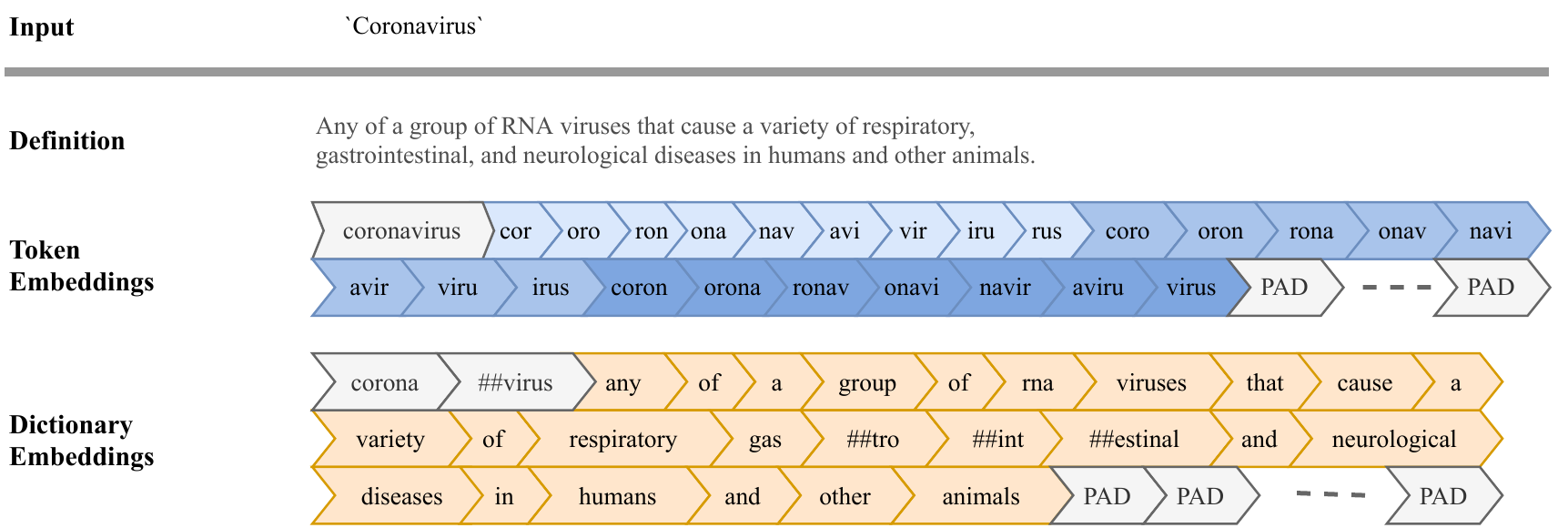}
    \caption{
    \textbf{Input sequence embeddings.}
    We use two encoding schemes to prepare input sequences for our models:  
    token embeddings (blue)
    and dictionary embeddings (orange)
    for our Token and Dictionary models, respectively.
    Given an input word (e.g., `coronavirus'), 
    we first break the input token into character-level $n$-grams ($n \in \{3, 4, 5\}$). 
    The resulting sequence of $n$-grams along with the original word at the beginning of the embeddings 
    are used in our Token model.
    Sequences shorter than a specified length are appended with PAD, a padding token ensuring a universal input size.
    For our Dictionary model, 
    we first look up a dictionary definition for the given input. 
    We then process the input word along with its definition into subwords using WordPiece~\cite{wu2016wordpiece}. 
    Uncommon and novel words are broken into subwords, 
    with double hashtags indicating that the given token is not a full word.
    }
    \label{fig:embeddings}
\end{figure*}

\begin{figure}[!tp]
    \centering
    \includegraphics[width=.86\columnwidth]{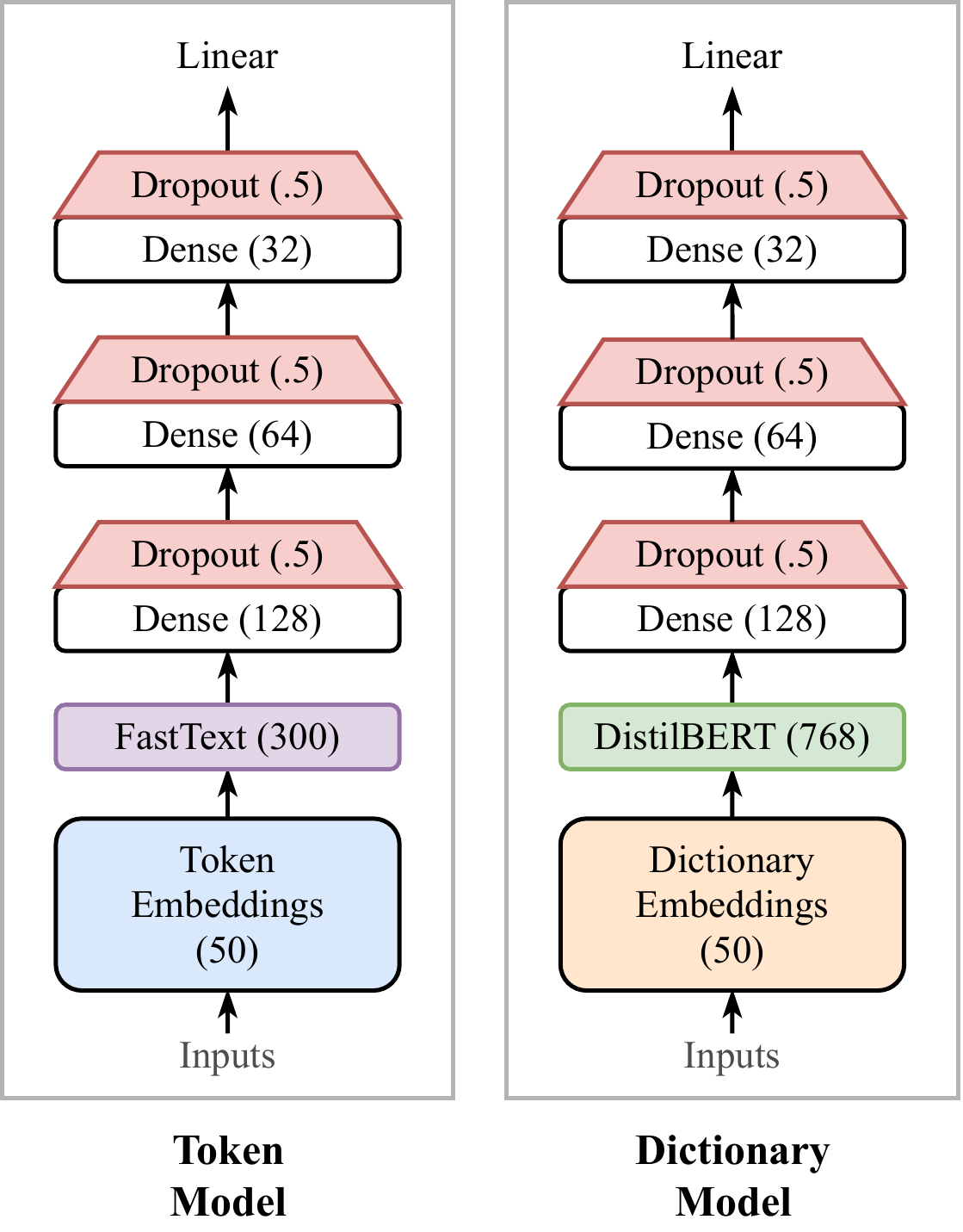}
    \caption{
    \textbf{Model architectures.}
    Our first model is a neural network
    initialized with pre-trained word embeddings
    to estimate happiness scores.
    Our second model,
    is a deep Transformer-based model that uses word definitions
    to estimate their sentiment scores.
    See Sec.~\ref{subsec:sentiment.token_model}
    and Sec.~\ref{subsec:sentiment.dictionary_model}
    for further technical details of each model, respectively.
    Note the Token model is considerably smaller with roughly 10 million trainable parameters 
    compared with the Dictionary model that has a little over 66 million parameters.
    }
    \label{fig:models}
\end{figure}

\subsection{Token Model}
\label{subsec:sentiment.token_model}

Our first model uses a neural network that learns
to map words from the labMT lexicon to their corresponding sentiment scores.
While still being able to learn a non-linear mapping between the words and their happiness scores, the model only considers the individual words as input---enriching its internal utility function with subword representations to estimate the happiness score.

The input word is first processed into a token
embedding---sequentially breaking each word into its equivalent character-level $n$-grams whereby 
$n \in \{3, 4, 5\}$
(see Fig.~\ref{fig:embeddings} for an illustration).
English words have an average length of 5 characters~\cite{miller1958length,mayzner1965tables},
which would yield 6 unique character-level $n$-grams given our tokenization scheme.
While we did try shorter and longer sequences,
we fix the length of the input sequence to a size of 50 and pad shorter sequences to ensure a universal input size.
We choose a longer sequence length to allow us to encode longer $n$-grams and rare words.  

We then pass the token embeddings to a 300-dimensional embedding layer.
We initialize the embedding layer with weights
trained with subword information on Common Crawl and Wikipedia using FastText~\cite{bojanowski2017enriching}.
In particular,
we use weights from a pre-trained model using CBOW
with character-level $n$-grams of length 5 and a window size of 5 and 10 (\url{https://fasttext.cc/docs/en/english-vectors.html}).

The output of the embedding layer is pooled down and passed
to a sequence of three dense layers of decreasing sizes: 128, 64, and 32, respectively.
We use a rectified linear activation function (ReLU) for all dense layers.
We also add a dropout layer after each dense layer,
with a 50\% dropout rate to add stochasticity to the model,
allowing for a simple estimate of uncertainty using the standard deviation of the network's predictions~\cite{srivastava2014dropout}.

We experimented with a few different layout configurations,
finding that making the network either wider or deeper has minimal effect on the network performance.
Therefore,
we choose to keep our model rather simple with roughly 10 million trainable parameters.
The output of the last dense layer is finally passed over to a single output layer with a linear activation function to regress a sentiment score between 1 and 9.
See Fig.~\ref{fig:models} for a simple diagram of the model architecture.

\subsection{Dictionary Model}
\label{subsec:sentiment.dictionary_model}

Historically, 
lexicon-based models have only considered simple statistical methods 
to estimate the emotional valence of words. 
Here, 
we try to bridge the connection between the conventional techniques among the community and recent advances in NLP.

For our second model, 
we use a contextualized Transformer-based language model 
to estimate the sentiment score for a given word based on its dictionary definition.
While still predicting scores for individual words, 
we now do so by augmenting each word with its expressed meaning(s) from a general dictionary. 
Given an input word, 
we look up its definition via a free online dictionary API available at \url{https://dictionaryapi.dev}.

The average length of definitions for the words found in labMT is roughly 38 words.
We choose a maximum definition length of 50 words---which covers the 75th percentile of that distribution---to ensure that words with multiple definitions are adequately represented. 
While increasing the sequence length beyond 50 did not improve our accuracy, it increases the model complexity slowing our training and inference time substantially. 
Therefore, we fix the length of word definitions to a maximum of 50 words.
We pad shorter sequences, and truncate words 51 and beyond to ensure a fixed input size.

We estimate the sentiment of each labMT word as follows. 
The word,
along with its definition, 
is processed into dictionary embeddings 
by breaking each word into subwords based on their frequency of usage using WordPiece~\cite{wu2016wordpiece}. 
This is a widely adopted tokenization technique
that breaks uncommon and novel words into subwords,
which reduces the vocabulary size of language models 
and enables them to handle OOV tokens. 
Other tokenization models will give similar results~\cite{kudo2018sentencepiece}.
We only use the word as input to our model for terms without definitions. 

In principle, 
the dictionary embeddings can be passed to a vanilla Transformer model 
(e.g., BERT~\cite{devlin2019bert}, XLNet~\cite{yang2019xlnet}).
However, 
we prefer more manageable models 
(i.e., smaller and faster) 
due to their efficiency while maintaining state-of-the-art results. 
We tried both ALBERT~\cite{lan2020albert} and DistilBERT~\cite{sanh2019distilbert}. 
Both models have equivalent performance on our task. 
The output of the model's pooling layer is passed to 
a sequence of three dense layers of decreasing sizes 
with dropout applied after each layer---similar to our approach in the Token model.
Finally, 
the output of the last dense layer is projected down to a single output value 
that servers as the sentiment score prediction.

The Token model is considerably lighter in terms of memory usage, 
and faster in terms of training and inference time than the Dictionary model. 
Our current configuration of the Token model results in roughly 10 million trainable parameters 
compared with the Dictionary model that has over 66 million parameters.

\section{Results and discussion}
\label{sec:sentiment.results}


\begin{table*}[!tp]
    \centering
    \begin{tabularx}{\textwidth}{l CCCCCC}
    \multicolumn{2}{l}{\textbf{Mean absolute error (MAE)}} & \multicolumn{5}{c}{\textbf{Percentiles}} \\ \hline
    \textbf{Model} & \textbf{Average} & \textbf{$25^{th}$} & \textbf{$50^{th}$} & \textbf{$75^{th}$} & \textbf{$85^{th}$} & \textbf{$95^{th}$} \\
    \hline\hline
    \multicolumn{7}{c}{Linear models} \\
    ElasticNet + \textit{Word2Vec}  & 0.81 & 0.82 & 0.81 & 0.82 & 0.82 & 0.83 \\
    ElasticNet + \textit{GloVe}     & 0.81 & 0.82 & 0.81 & 0.82 & 0.82 & 0.82 \\
    ElasticNet + \textit{FastText}  & 0.81 & 0.82 & 0.81 & 0.82 & 0.82 & 0.82 \\
    LASSO + \textit{Word2Vec}       & 0.81 & 0.81 & 0.81 & 0.81 & 0.82 & 0.83 \\
    LASSO + \textit{GloVe}          & 0.81 & 0.81 & 0.82 & 0.82 & 0.82 & 0.82 \\
    LASSO + \textit{FastText}       & 0.81 & 0.80 & 0.81 & 0.81 & 0.81 & 0.82 \\ 
    Ridge + \textit{Word2Vec}       & 0.73 & 0.73 & 0.73 & 0.74 & 0.74 & 0.75 \\ 
    Ridge + \textit{GloVe}          & 0.75 & 0.74 & 0.75 & 0.75 & 0.77 & 0.79 \\
    Ridge + \textit{FastText}       & 0.73 & 0.73 & 0.73 & 0.74 & 0.74 & 0.74 \\ \hline
    \multicolumn{7}{c}{Random forest (RF) models} \\
    RF + \textit{Word2Vec}          & 0.69 & 0.69 & 0.70 & 0.70 & 0.71 & 0.78 \\
    RF + \textit{GloVe}             & 0.70 & 0.70 & 0.70 & 0.71 & 0.71 & 0.71 \\
    RF + \textit{FastText}          & 0.68 & 0.67 & 0.68 & 0.68 & 0.68 & 0.69 \\ \hline
    \multicolumn{7}{c}{Support vector regressor (SVR) models} \\
    SVR + \textit{Word2Vec}         & 0.65 & 0.65 & 0.65 & 0.66 & 0.66 & 0.67 \\
    SVR + \textit{GloVe}            & 0.67 & 0.68 & 0.67 & 0.66 & 0.68 & 0.69 \\
    SVR + \textit{FastText}         & 0.64 & 0.64 & 0.64 & 0.65 & 0.66 & 0.66 \\ \hline
    \multicolumn{7}{c}{Proposed models} \\
    \textbf{Token Model (single)}               & 0.62 & 0.60 & 0.61 & 0.64 & 0.65 &  0.66 \\
    \textbf{Token Model (ensemble)}             & 0.57 & 0.29 & 0.44 & 0.66 & 0.72 & 0.77 \\ 
    \textbf{Dictionary Model (single)}          & 0.50 &  0.49 & 0.50 & 0.51 &  \textbf{0.51} & \textbf{0.52} \\
    \textbf{Dictionary Model (ensemble)}        & \textbf{0.45}  & \textbf{0.15} & \textbf{0.31} & \textbf{0.40} & 0.52 & 0.59 \\ \hline
    Human ratings (standard deviation $\sigma$) & 1.38 & 1.18 & 1.36 & 1.56 & 1.69 & 1.90 \\
    Human ratings (variance $\sigma^2$)         & 1.99 & 1.39 & 1.85 & 2.43 & 2.86 & 3.61\\ 
    \end{tabularx}
    \caption{
        Summary statistics of the testing subset comparing our models to the annotated ratings reported in labMT. 
        Each word in the labMT lexicon is scored by 50 distinct individuals 
        and the final happiness score is the unweighted mean of their scores~\cite{dodds2015human}. 
        We report the standard deviation and variance of the ratings 
        as a baseline to assess the human's confidence in the reported scores.   
        Comparing our predictions with the annotations crowdsourced via AMT, 
        our mean absolute errors are on par with the variance observe in the human annotated labMT scores.
        In addition to our proposed models, 
        we also evaluate three groups of baseline models based on linear regression, random forests, and support vector machines.
        We trained and evaluated each baseline model over 10 trials using one of three pre-trained embeddings as the primary input: 
        \texttt{word2vec-google-news-300}~\cite{mikolov2013efficient}, \texttt{glove-wiki-gigaword-300}~\cite{pennington2014glove},
        \texttt{fasttext-wiki-news-subwords-300}~\cite{bojanowski2017enriching}.
        Our baseline models are similar to those seen in~\citet{amir_INESCID_2015} and~\citet{li_Inferring_2017}.
    }
    \label{tab:model_maes}
\end{table*}

\begin{figure}[!tp]
    \centering
    \includegraphics[width=\columnwidth]{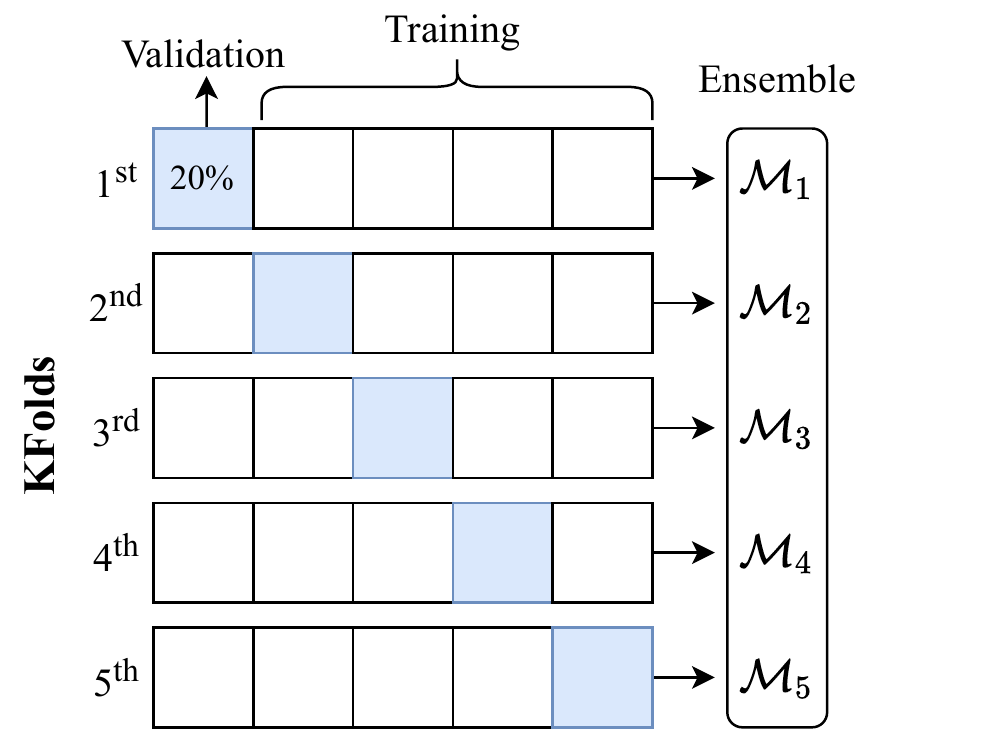}
    \caption{
    \textbf{Ensemble learning and $k$-fold cross-validation.}
    Using an 80/20 split for training/validation, 
    we train our models for a maximum of 500 epochs per fold for a total of 5 folds. 
    We use the model trained from each fold to build an ensemble 
    because the average performance of an ensemble is less biased and better than the individual models. 
    }
    \label{fig:ensemble}
\end{figure}

\begin{figure*}[!tp]
    \centering
    \includegraphics[width=\textwidth]{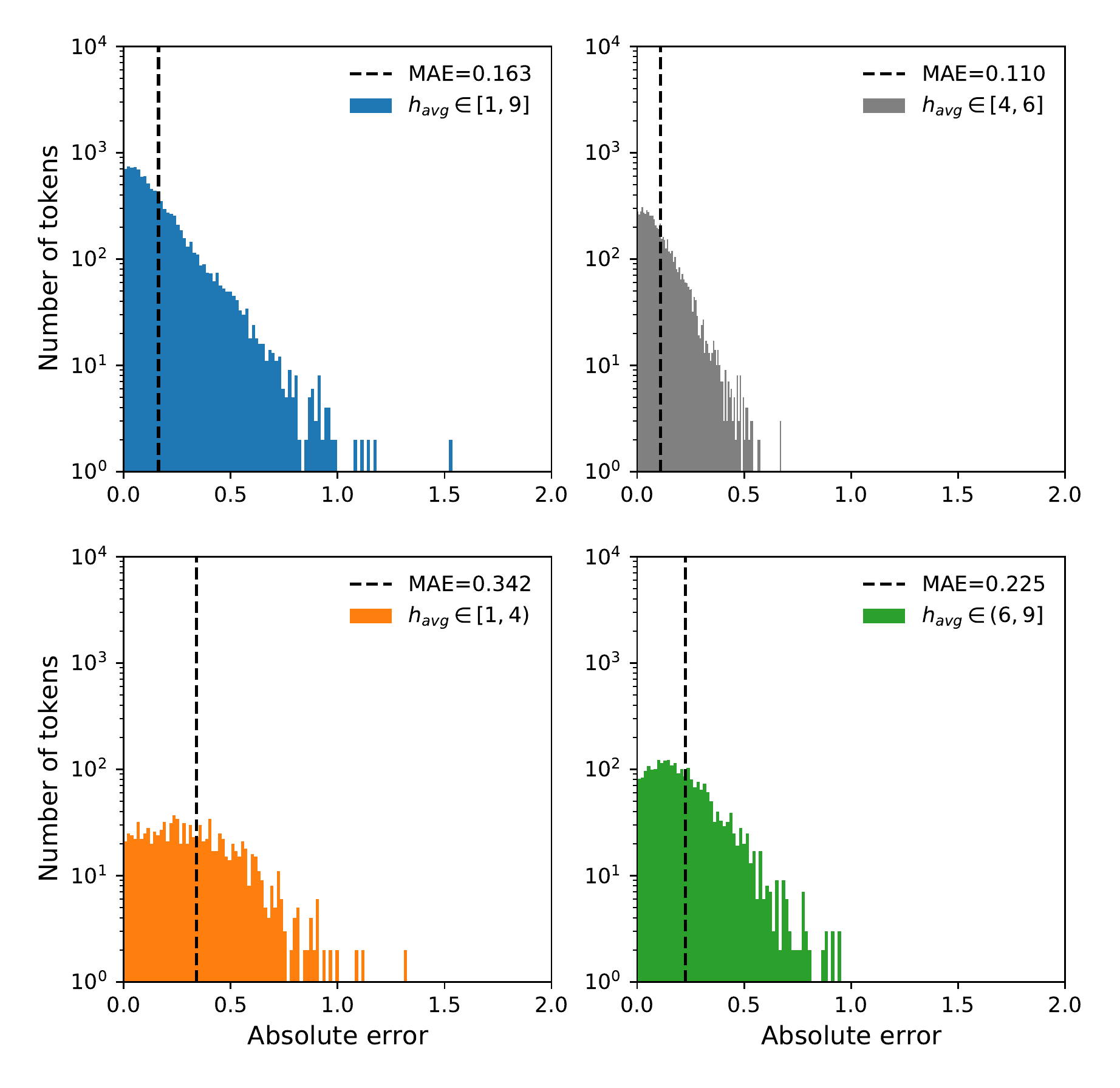}
    \caption{
    \textbf{Error distributions for the Token model.}
    We display mean absolute errors for predictions using the Token model on all words in labMT.
    We arrange the happiness scores into three groups:
    negative ($h_{avg} \in [1,4)$, orange),
    neutral ($h_{avg} \in [4,6]$, grey),
    and positive ($h_{avg} \in (6,9]$, green).
    Most words have an MAE less than 1 with the exception of a few outliers.  
    We see a relatively higher MAE for negative and positive terms compared to neutral expressions.
    }
    \label{fig:labmt_dists_toekn}
\end{figure*}

\begin{figure*}[!tp]
    \centering
    \includegraphics[width=\textwidth]{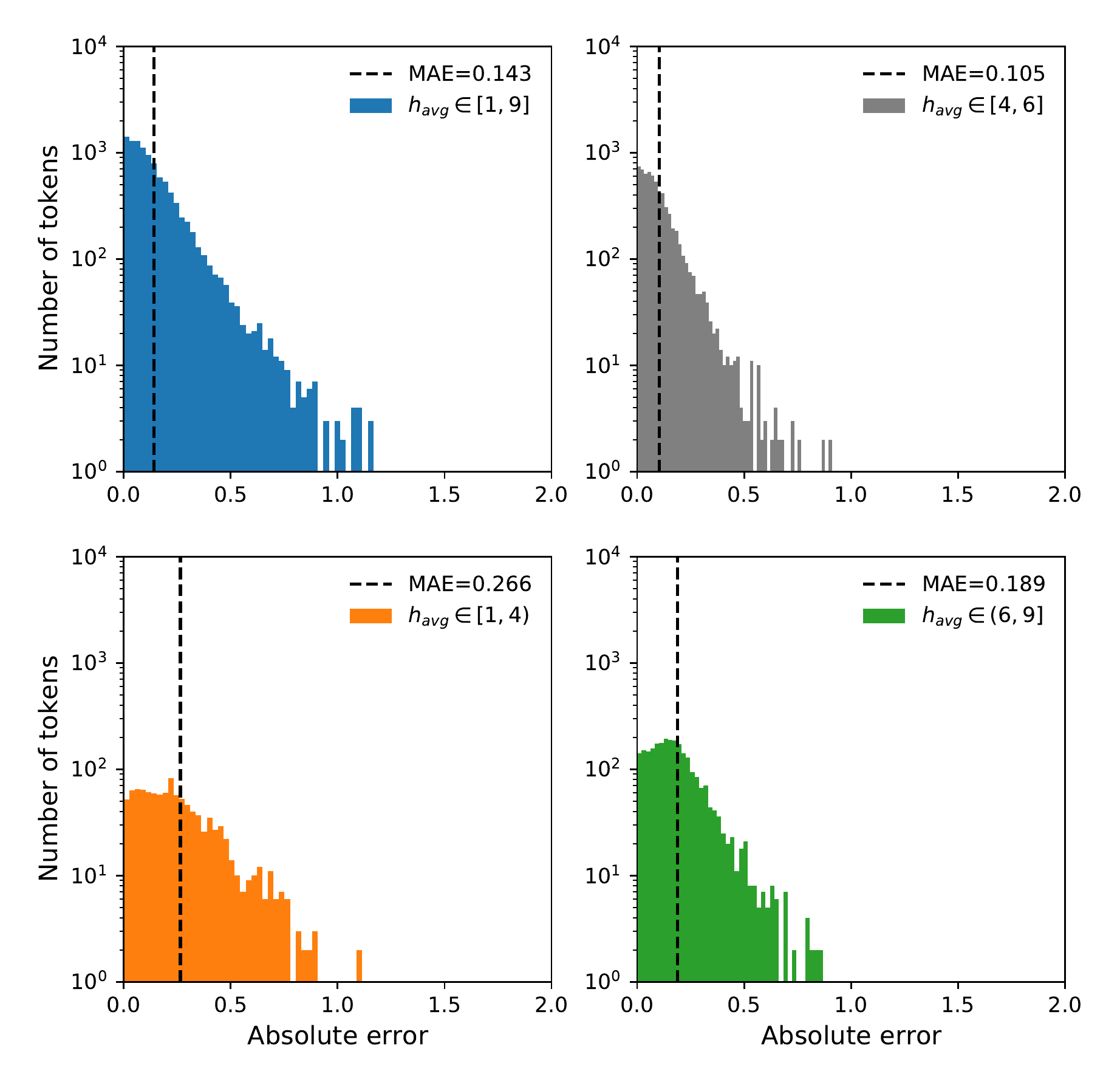}
    \caption{
    \textbf{Error distributions for the Dictionary model.}
    We display mean absolute errors for predictions using the Dictionary model on all words in labMT.
    Again, we categorize the happiness scores into three groups:
    negative ($h_{avg} \in [1,4)$, orange),
    neutral ($h_{avg} \in [4,6]$, grey),
    and positive ($h_{avg} \in (6,9]$, green).
    Similar to the Token model, 
    most words have an MAE less than 1 with the exception of a few outliers.  
    While the Dictionary model outperforms the Token model, we still observe a higher MAE for negative and positive terms compared to neutral expressions. 
    }
    \label{fig:labmt_dists_dict}
\end{figure*}

\begin{figure*}[!tp]
    \centering
    \includegraphics[width=\textwidth]{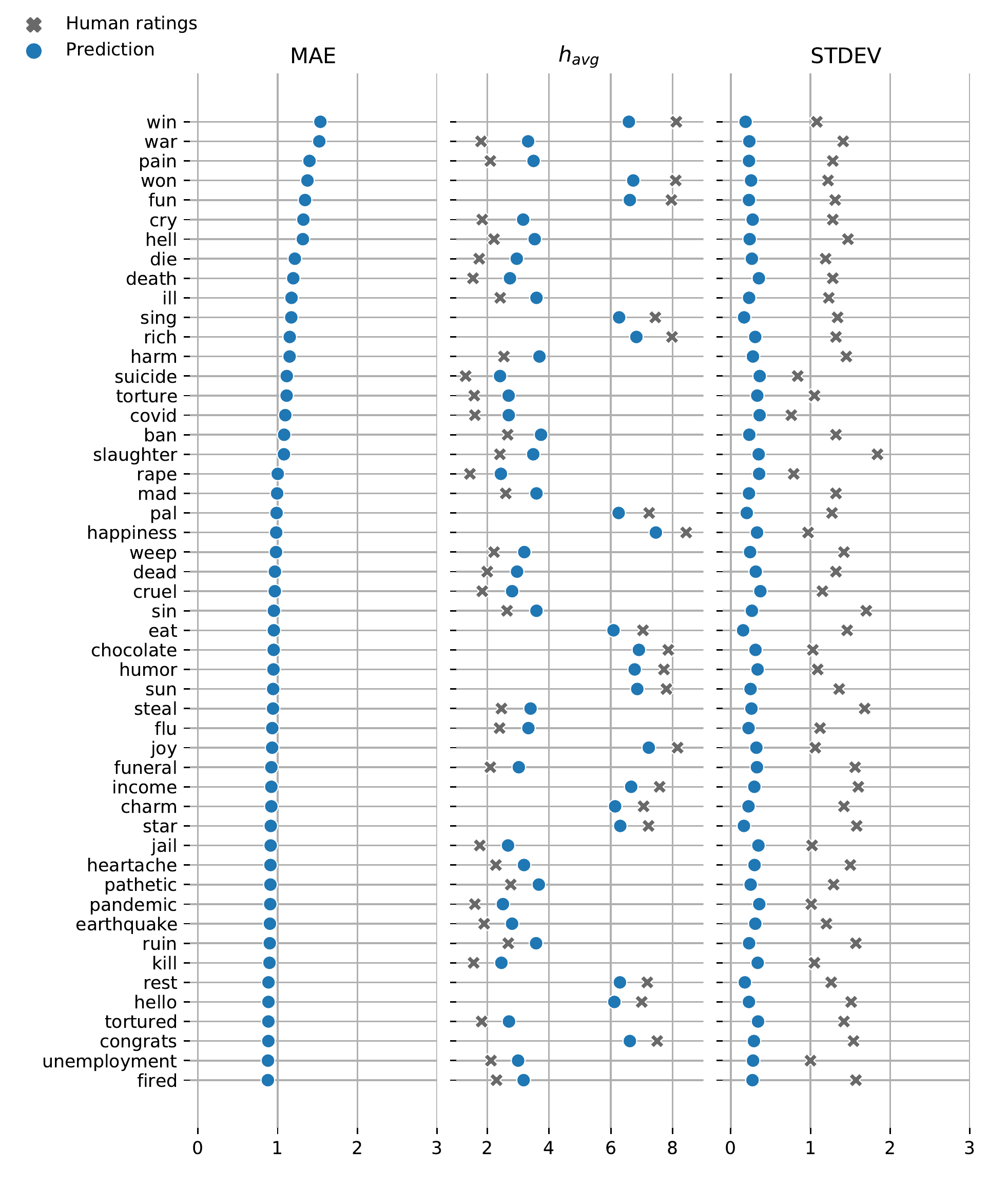}
    \caption{
    \textbf{Token model: Top-50 words with the highest mean absolute error.}
    Model predictions are shown in blue and the crowdsourced annotations are displayed in grey. 
    While still maintaining relatively low MAE, most of our predictions are 
    conservative---marginally underestimating words with extremely high happiness scores,
    and overestimating words with low happiness scores. 
    }
    \label{fig:happiness_conf_token}
\end{figure*}

\begin{figure*}[!tp]
    \centering
    \includegraphics[width=\textwidth]{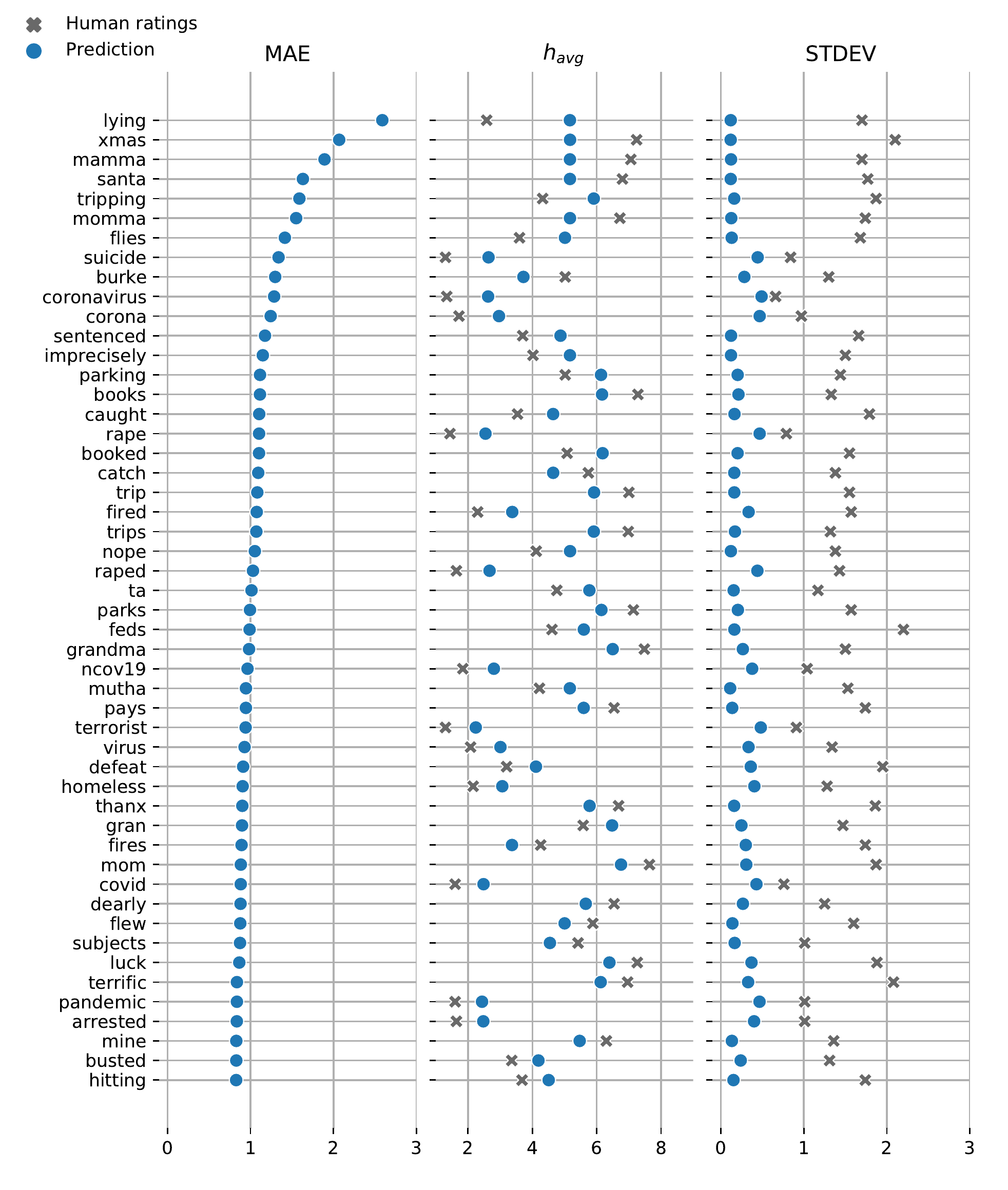}
    \caption{
    \textbf{Dictionary model: Top-50 words with the highest mean absolute error.}
    Model predictions are shown in blue and the crowdsourced annotations are displayed in grey. 
    Note, the vast majority of words with relatively high MAE also have high standard deviations of AMT ratings.
    Words that have multiple definitions will have a neutral score (e.g., lying).  
    A neutral happiness score is also often predicted for words because we are unable to obtain good definitions for them to use as input.
    Although we have definitions for most words in our dataset,
    we still have a little over 1500 words with missing definitions.
    Most of these words are
    names (e.g., `Burke'), 
    and slang (e.g., `xmas', and `ta').
    }
    \label{fig:happiness_conf_dict}
\end{figure*}

\subsection{Ensemble learning and $k$-fold cross-validation}

In the deep learning community, particularly in the NLP domain, 
it is common to scale up the number of parameters in successful models to eke out additional performance gains.
The effectiveness of this approach tends to be correlated with the amount of training data available (i.e., larger models are more effective when trained on larger data sets).
With the limited size of our training set, we needed alternative techniques to increase the performance of our models.
Ensemble learning is a widely known and adopted family of methods 
in which the average performance of an ensemble is shown to be 
both less biased and better than the individual models~\cite{hansen1990neural,krogh1994neural}.

First, 
we randomly subsample our dataset, taking a 20\% subset as our holdout set for testing. 
Using a 5-fold cross-validation strategy, 
we break the remaining samples into 5 distinct subsets using a 80/20 split for training/validation. 
We train one model per fold for a maximum of 500 epochs each,
and combine the 5 trained models to form an ensemble.
While there are many gradient descent optimization algorithms, 
we use Adam~\cite{kingma2015adam} as a popular and well-established optimizer, 
keeping its default configuration and setting our initial learning rate to 0.001.
In Fig.~\ref{fig:ensemble},
we show a breakdown of our ensemble pipeline 
whereby the blue squares highlight the validation subset for each fold. 
Note, 
the holdout set is removed before training the ensemble and is only used for testing a complete ensemble.

To estimate the happiness score for a given word, 
we take a Monte Carlo approach by sampling 100 predictions per model in the ensemble. 
We use the training setting for the dropout layers in each model, 
rather than the test time averaging that is commonly used, so that these predictions are heterogeneous. 
The mean over these predictions becomes the proposed happiness score, 
while the standard deviation serves as an estimate of model uncertainty~\cite{gal2016dropout}. 
Providing a point estimate along with an uncertainty band 
allows us to compare and contrast the level of model uncertainty in our ensembles 
with the uncertainty observed between human annotators.

\subsection{Comparison with other methods and human annotators} 

Although both of our proposed strategies---namely using character-level $n$-grams and word definitions---performed well, the Dictionary model outperforms the Token model. 
To evaluate our models we train 10 replicates each and then investigate error distributions obtained using the test set.
We report the mean absolute error (MAE) as an estimate of overall performance, 
along with a selection of percentiles to compare tail behavior across models.
Each of these statistics are averaged over the 10 replicates.
This process provides us with a strong estimate of the generalization performance for our proposed models.

Table~\ref{tab:model_maes} summarizes the results of this evaluation process for our proposed models and ensembles.
We provide baseline comparisons to models from previous work~\cite{amir_INESCID_2015,li_Inferring_2017}, 
including popular linear models, random forests, and support vector machines trained with three different flavors of word embeddings: 
Word2Vec~\cite{mikolov2013efficient}, GloVe~\cite{pennington2014glove}, and FastText~\cite{bojanowski2017enriching}.
These results indicate that our Token model outperforms all prior baselines, our Dictionary model outperforms our Token model, and both of our proposed models benefited from ensemble learning.
Though the ensembles outperformed the individual models in both cases, it is interesting to note that they also had longer tails for their error distributions.

We further examine the error distributions 
to investigate if the models have a bias towards high or low happiness scores. 
In Figs.~\ref{fig:labmt_dists_toekn} and~\ref{fig:labmt_dists_dict}, 
we display a breakdown of our MAE distributions for the Token and Dictionary models, respectively. 
For ease of interpretation and visualization,  
we categorize the happiness scores into three groups: 
negative ($h_{avg} \in [1,4)$), 
neutral ($h_{avg} \in [4,6]$), 
and positive ($h_{avg} \in (6,9]$). 
While the distributions show our models operate well on all words, 
particularly neutral expressions, 
we note a relatively higher MAE for negative words, 
whereby our predictions to these terms are more positive than the annotations.


We also compare our predictions to the ground-truth ratings, 
examining the degree to which the models either overshoot 
or undershoot the happiness scores crowdsourced via AMT.
Words in the labMT lexicon were scored by taking the average happiness score of 
distinct evaluations from 50 different individuals (see Table~S2~\cite{dodds2015human}). 
Since the variance of human ratings and our model MAEs are on the same scale, 
we can use the observed average variance of the ratings (1.17) 
as a baseline to assess rater confidence in the reported scores.   
Comparing our models to that baseline, 
we note that all models offer consistent predictions 
with similar expectations to a random and reliable reviewer from AMT. 
See Table~\ref{tab:model_maes} for further statistical details.

In Figs.~\ref{fig:happiness_conf_token} and~\ref{fig:happiness_conf_dict}, 
we display the top-50 words with the highest mean absolute error for the Token and Dictionary models, respectively.
While the models always predict the right emotional attitude outlining each word based on its lexical polarity, 
they bias toward neutral by undershooting scores for happy words, 
and overshooting scores for sad expressions.

One possible explanation of this systematic behavior 
is the lack of words with extreme happiness scores in the labMT lexicon. 
It is possible to train models with a smaller but balanced subset of the dataset to overcome that challenge. 
Doing so, 
however, 
would reduce the size of training/validation samples substantially. 
Still, 
our margin of error is relatively low compared to human ratings. 
Future investigations may test and improve the models by examining larger sentiment lexicons.


Another key factor that plays a big role in our prediction error is obtaining good word definitions, 
or the lack thereof, 
to use as input for our Dictionary model.
Surprisingly, 
outsourcing definitions from online dictionaries for a large set of words is rather challenging, 
especially if you opt-out of reliable but paid services.    
In our work, 
we choose not to use an urban dictionary or any services with paid APIs.   
We use a free online dictionary API that is available at 
\url{https://dictionaryapi.dev}. 

While we do have definitions for most words in our dataset, 
a total of 1518 words have missing definitions. 
Most of these words are 
names, 
abbreviations, 
and slang terms (e.g., `xams', `foto', `nvm', and `lmao').  
Words with multiple definitions can also cancel each other's score (e.g., `lying').


Notably, 
the vast majority of words with high MAE also have high AMT standard deviations.  
To further investigate prediction accuracy, 
we examine the overlap between the predictions and human ratings. 
In particular, 
we compute the intersection over union (IOU) between the predicted happiness score 
$h'_{avg} \pm \sigma'$,
and the corresponding value from the annotated ratings 
$h_{avg} \pm \sigma$. 

The Token model underestimates the happiness score for 
`win'---the only word with a prediction that falls outside the range of human annotated happiness scores. 
The remaining predicted happiness scores fall well within the range of scores crowdsourced via AMT. 
Similarly, 
the Dictionary model slightly underestimates the happiness scores for `mamma' 
while overestimating the scores for `lying', and `coronavirus'.

\section{Concluding remarks}
\label{sec:sentiment.concludingremarks}


As the growing demand for sentiment-aware intelligent systems increases, 
we will continue to see improvements to both lexicon-based models and contextual language models. 
While contextualized models are suitable for a wide set of applications, 
lexicon-based models are used by 
computational linguists, 
journalists, 
and data scientists 
who are interested in studying how individual words contribute to sentiment trends.

Sentiment lexicons, 
however, 
have to be updated periodically to support new words and expressions that were not considered when the dictionaries were assembled.
In this paper, we proposed two models for predicting sentiment scores
to augment semantic dictionaries using word embeddings and pre-trained large language models.
Our first model establishes a baseline using a neural network initialized with pre-trained word embeddings, 
while our second model features a deep Transformer-based network 
that brings into play word definitions to estimate their lexical polarity.
Our results and evaluation of both models demonstrate human-level performance on a state-of-the-art human annotated list of words.


Although both models can predict scores for novel words, 
we acknowledge a few shortcomings.
Our Token model relies on subword information to estimate a happiness score for any given word. 
For example, 
using subwords for `coronavirus' yields a good estimate given that it contains `virus'.
By contrast, 
parsing character-level $n$-grams for other words (e.g., `covid') may not reveal any further information.
We can overcome that hurdle by using the word definition as input to our Dictionary model to gauge its happiness score. 
Words, 
however, 
often have different meanings based on context. 
Finding good definitions may be challenging, especially for slang, informal expressions, and abbreviations. 
We recommend using the Dictionary model whenever it is possible to outsource a good definition of the word.


A natural next step would be to develop similar models for other languages, 
for example by building a model for each language,
or a multilingual model.
Fortunately, 
FastText~\cite{bojanowski2017enriching} provides pre-trained word embeddings for over 100 languages. 
Therefore, it is easy to upgrade the Token model to support other languages.
Updating the Dictionary model is also a straightforward task 
by simply adopting a multilingual Transformer-based model pre-trained with several languages 
(e.g., Multilingual BERT~\cite{devlin2019bert}).
We caution against translating words and using the same English scores 
because most words do not have a one-to-one mapping into other languages, 
and are often used to express different meanings by the native speakers of any given language~\cite{dodds2015human}.


Another vast space of improvements would be to adopt our proposed strategies
to develop prediction models for other semantic dictionaries. 
Researchers can further fine-tune these models to predict other sentiment scores.
For example, 
the happiness scores in the labMT~\cite{dodds2015human} dataset are closely aligned with 
the valence scores in the NRC-VAD lexicon~\cite{mohammad2018obtaining}.
We envision future work developing similar models to predict other semantic differentials 
such as arousal and dominance~\cite{mohammad2018obtaining}, 
EPA~\cite{osgood1962studies},
and SocialSent~\cite{hamilton2016inducing}.
Our primary goal is to provide an easy and robust method to augment semantic dictionaries 
to empower researchers to maintain and expand them at a relatively low cost 
using today's state-of-the-art NLP methods.


\acknowledgments
We are grateful for the computing resources provided by the Vermont Advanced Computing Core 
and financial support from Google and the Massachusetts Mutual Life Insurance Company.
We thank Anne Marie Stupinski and Julia Zimmerman for their insightful discussion and suggestions.
Computations were performed on the Vermont Advanced Computing Core supported in part by NSF award No. OAC-1827314.

\bibliography{\filenamebase}

\bibliographystyle{abbrvnat}

\clearpage


\end{document}